\documentclass{IEEEtran4PSCC}
\pdfoutput=1
\ifCLASSINFOpdf
   \usepackage[pdftex]{graphicx}
\else
   \usepackage[dvips]{graphicx}
\fi

\usepackage[cmex10]{amsmath}
\usepackage[normalem]{ulem}
\usepackage{amsfonts}
\usepackage{soul}
\usepackage{float}
\usepackage[noabbrev]{cleveref}
\usepackage{url}
\usepackage{calrsfs}
\usepackage{textcomp} 
\usepackage{multirow}
\usepackage[monochrome]{color}

\makeatletter
\let\old@ps@headings\ps@headings
\let\old@ps@IEEEtitlepagestyle\ps@IEEEtitlepagestyle
\def\psccfooter#1{%
    \def\ps@headings{%
        \old@ps@headings%
        \def\@oddfoot{\strut\hfill#1\hfill\strut}%
        \def\@evenfoot{\strut\hfill#1\hfill\strut}%
    }%
    \def\ps@IEEEtitlepagestyle{%
        \old@ps@IEEEtitlepagestyle%
        \def\@oddfoot{\strut\hfill#1\hfill\strut}%
        \def\@evenfoot{\strut\hfill#1\hfill\strut}%
    }%
    \ps@headings%
}
\makeatother

\psccfooter{%
        \parbox{\textwidth}{\hrulefill \\ \small{23rd Power Systems Computation Conference} \hfill \begin{minipage}{0.2\textwidth}\centering \vspace*{4pt} \includegraphics[scale=0.06]{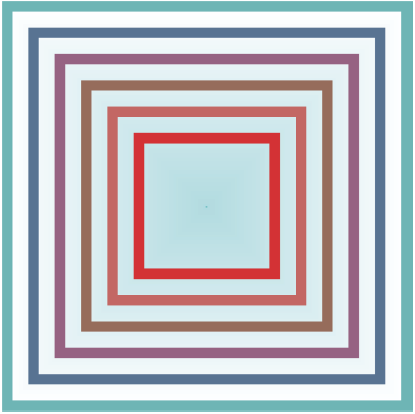}\\\small{PSCC 2024} \end{minipage} \hfill \small{Paris, France --- June 4 -- 7, 2024}}%
}

\begin{document}
\title{Interpretable Short-Term Load Forecasting via Multi-Scale Temporal Decomposition}

%% To specify the authors when (number of affiliations <= 2)
%\author{
%\IEEEauthorblockN{Author n.1 Name per Affiliation A\\ Author n.2 Name per Affiliation A}
%\IEEEauthorblockA{(Affiliation A) Department Name of Organization \\
%Name of the organization, acronyms acceptable\\
%City, Country\\
%\{email author n.1, email author n.2\}@domain (if desired)}
%\and
%\IEEEauthorblockN{Author n.1 Name per Affiliation B\\ Author n.2 Name per Affiliation B}
%\IEEEauthorblockA{(Affiliation B) Department Name of Organization \\
%Name of the organization, acronyms acceptable\\
%City, Country\\
%\{email author n.1, email author n.2\}@domain (if desired)}
%}

%% To specify the authors when (number of affiliations > 2)
\author{
 \IEEEauthorblockN{Yuqi Jiang, Yan Li}
  \IEEEauthorblockA{\textit{Department of Electrical and Computer Engineering} \\
    \textit{Pennsylvania State University, PA, USA}
 \\
    \{yzj5282, yql5925\}@psu.edu}
  \and
  \IEEEauthorblockN{ Yize Chen}
  \IEEEauthorblockA{\textit{AI Thrust, Information Hub} \\
    \textit{Hong Kong University of Science and Technology, China}\\
    yizechen@ust.hk}
 }

\maketitle

\begin{abstract}
Rapid progress in machine learning and deep learning has enabled a wide range of applications in the electricity load forecasting of power systems, for instance, univariate and multivariate short-term load forecasting. Though the strong capabilities of learning the non-linearity of the load patterns and the high prediction accuracy have been achieved, the interpretability of typical deep learning models for electricity load forecasting is less studied. This paper proposes an interpretable deep learning method, which learns a linear combination of neural networks that each attends to an input time feature. We also proposed a multi-scale time series decomposition method to deal with the complex time patterns. Case studies have been carried out on the Belgium central grid load dataset and the proposed model demonstrated better accuracy compared to the frequently applied baseline model. \textcolor{blue}{Specifically, the proposed multi-scale temporal decomposition achieves the best MSE, MAE and RMSE of 0.52, 0.57 and 0.72 respectively. As for interpretability, on one hand, the proposed method displays generalization capability. On the other hand, it can demonstrate not only the feature but also the temporal interpretability compared to other baseline methods}. Besides, the global time feature interpretabilities are also obtained. Obtaining global feature interpretabilities allows us to catch the overall patterns, trends, and cyclicality in load data while also revealing the significance of various time-related features in forming the final outputs. 
\end{abstract}


\section{Introduction}
As a cornerstone of power system management, electricity load forecasting plays a pivotal role in ensuring the balancing, efficiency, and cost-effectiveness of power grid operations. It serves as one of the foundation tools for the optimization of power generation and dispatch. Of particular significance is short-term load forecasting (STLF), for instance, forecasting the day-ahead or hour-ahead demand vectors for a region of interest~\cite {gross1987short}. Such task presents unique and immediate benefits for grid operation. It assists in the crucial task of balancing supply and demand on an hourly or daily basis, facilitating responsive and adaptive power distribution. In recent years, machine learning (ML)-based methods have emerged as prominent tools, largely due to their ability to learn complex non-linear patterns, handle diverse datasets, and catch long-term temporal dependencies. 

While deep learning models have greatly enhanced the accuracy of short-term load forecasting, one persisting concern is their 'black box' nature \cite{castelvecchi2016can, chen2018machine, chen2019exploiting}. Given input such as electricity load and other auxiliary feature data (temperature, dew, etc.), these models can make highly accurate predictions, but their internal decision-making processes remain unclear. \textcolor{blue}{For power engineers, apart from trusting individual predictions made by the machine learning models, there is also a need to both understand the contributing factors of certain load behaviors and to evaluate the model comprehensively before deploying it ``in the wild"~\cite{zhou2022robust, liu2022learning}.} The inherent complexity, highly nonlinear, and complicated functions of these deep learning models often hinder them from providing clear data insights. Such insights can help understand patterns and trends in the historical load series. This includes identifying the trend and cyclical patterns that are key factors for future load forecasts~\cite{almeshaiei2011methodology}. Besides, data insights can help improve the accuracy of load forecasts. By understanding how input features affect future loads, the model settings such as feature representation and model size can be optimized. Though modern machine learning models excel at making predictions, yet the insights that could be derived from these findings remain unsolved, since most of them do not provide any correlation or importance analysis between outputs and inputs. 

The interpretable deep learning models developed for time series problems can be prescribed as ante-hoc and post-hoc interpretability based on learning phase~\cite{rojat2021explainable}. Ante-hoc methods incorporate interpretability into the models during the learning phase. Among them, deep Temporal Fusion Transformer (TFT) is a notable example, where interpretable self-attention is developed to incur model explanations. Yet in TFT the output forecasts need to be explained by all model weights rather than self-attention weights~\cite{jain2019attention}, making it less practical. TFT and other ante-hoc methods require that all components within the model are interpretable, which limits the design choices of forecasting models. Post-hoc interpretability refers to the process of interpreting a model after it has already been trained. Feature importance is one of the most important techniques, which involves identifying the important features accounting for the model outputs. Paper \cite{gurses2022introducing} introduces the saliency map to visually illustrate the relationships between the input features and the output short-term load forecasts. However, the generated saliency map is not robust to the shift of forecasting time. Post-hoc interpretability methods often better involve trade-offs between interpretability and accuracy. 

There are also distinctions between global and local interpretabilities~\cite{burkart2021survey}. Local interpretability concentrates on understanding individual predictions given a specific input. Global interpretability often involves understanding which features are most influential for the model across all instances \cite{gurses2022introducing}. This kind of interpretation is especially useful for getting a high-level understanding of a model's behavior through features, such as understanding the correlation between the model outputs and the input features (Weather, Humidity, etc.). In addition, explanations of the interpretable models should be easy to understand. For load forecasting tasks, it is important to let system operators or engineers quickly identify the most important factor that impacts the forecasting results, while understanding the temporal correlations across features and electricity load time series.

% \begin{figure*}
%     \centering
%     \includegraphics[scale=0.467]{Table_Ref.png}
%     \caption{The detailed process of cosine similarity. We use each blue rectangle to denote one series. }
%     \label{cosine}
% \end{figure*}

As for time series forecasting tasks especially electricity load forecasting, temporal interpretability is crucial for it provides insights into the inner patterns of the load data. Several techniques are also proposed for the temporal pattern analysis in load forecasting tasks. Temporal patterns in time series analysis refer to identifiable patterns or structures, which is crucial for various tasks, e.g.,  time series forecasting and anomaly detection. Such analysis includes analyzing the temporal components and the stationarity of the time series data~\cite{brockwell2002introduction, enders2008applied}. Apart from the temporal feature interpretability, we also attach importance to the explainability of auxiliary features (temperature, calendar, etc.). Auxiliary features can have significant impacts on short-term electricity consumption patterns. For instance, temperature fluctuations can drastically influence energy demands due to heating and cooling needs, and the distinction between weekdays and weekends can delineate consumption patterns attributable to human routines.

To tackle the temporal non-linearities, General Additive Models (GAM) are often used to learn the non-linearities\cite{chang2021node}. GAM incorporates decomposition into the time series and applies the smooth function to forecast the decomposed components. To realize the task of electricity load forecasting, \cite{obst2021adaptive} applies GAM to French load data during the COVID-19 lockdown. However, the proposed GAM method in \cite{obst2021adaptive} does not take auxiliary features into account, which limits its applicability for load forecasting tasks. Furthermore, paper \cite{agarwal2021neural} proposes a Neural Additive Model (NAM), which applies neural networks (NN) to replace the canonical smooth function in order to obtain better representations. Each neural network will attain a forecasting feature. At the output layer, NAM utilizes a linear layer to learn a linear combination of the outputs learned by all the neural networks. Though NAM performs generally better than the canonical GAM on many datasets, it does not include a temporal decomposition and has not been applied to load forecasting tasks. \textcolor{blue}{Recent papers \cite{zhang2023accurate, han2022efficient} also incorporate decomposition by utilizing CNN to analyze the temporal patterns and auxiliary features are also pre-processed. Inspired by the interpretable nature of both NAM and GAM and the recent decomposition applications in temporal pattern analysis, we aim to propose a method that \emph{incorporates both temporal and auxiliary feature interpretability}.}  %Besides, no tests on electricity load forecasting are carried out by NAM. 

\textcolor{blue}{In this paper, a novel post-hoc global interpretability method is proposed, and the main contribution is as follows:
\begin{itemize}
    \item The proposed method can be applied to  both continuous and categorical features. The flexible structure of the proposed model can be utilized for other datasets with various temporal patterns by adjusting the number of decomposition kernels and the kernel sizes. 
    \item The proposed method can find interpretability of the load data itself (temporal features) and the common auxiliary forecasting feature, while previous additive based SOTA model NAM can only provide feature interpretability.
    \item In the experiments on load data from Belgium and Australia which have distinct temporal and weather patterns, it can be demonstrated that the proposed method not only can outperform other forecasting modules in terms of forecasting accuracy, but also provide physically consistent interpretations.
\end{itemize}}

%In this paper, a novel post-hoc global interpretability method is proposed, which can be applied to datasets with different patterns and both continuous and categorical features.  We also propose to find interpretability of the load data itself (temporal features) and the common auxiliary forecasting feature. In the experiments, we test our model can be readily applied to load data from Belgium and Australia, which have different temporal and weather patterns. Comparison to baseline load forecasting models demonstrate that our method not only can outperform other forecasting modules in terms of forecasting accuracy, but also provide physically consistent  interpretations. The flexible structure of our proposed model can be utilized for other datasets with various temporal patterns by adjusting the number of decomposition kernels and the kernel sizes. 

\textcolor{blue}{The remainder of this paper is as follows: Section \ref{model} describes the main forecasting task and the overall framework of the proposed model. Section \ref{decomposition} introduces the detailed decomposition operation, which is a crucial part of the proposed method. In Section \ref{exp} the numerical studies are carried out to verify the effectiveness of the proposed method in terms of the forecasting accuracy and the interpretabilities. Section \ref{conclusion} summarizes this paper. }

\begin{figure*}[t]
    \centering
    \includegraphics[width=1.0\textwidth]{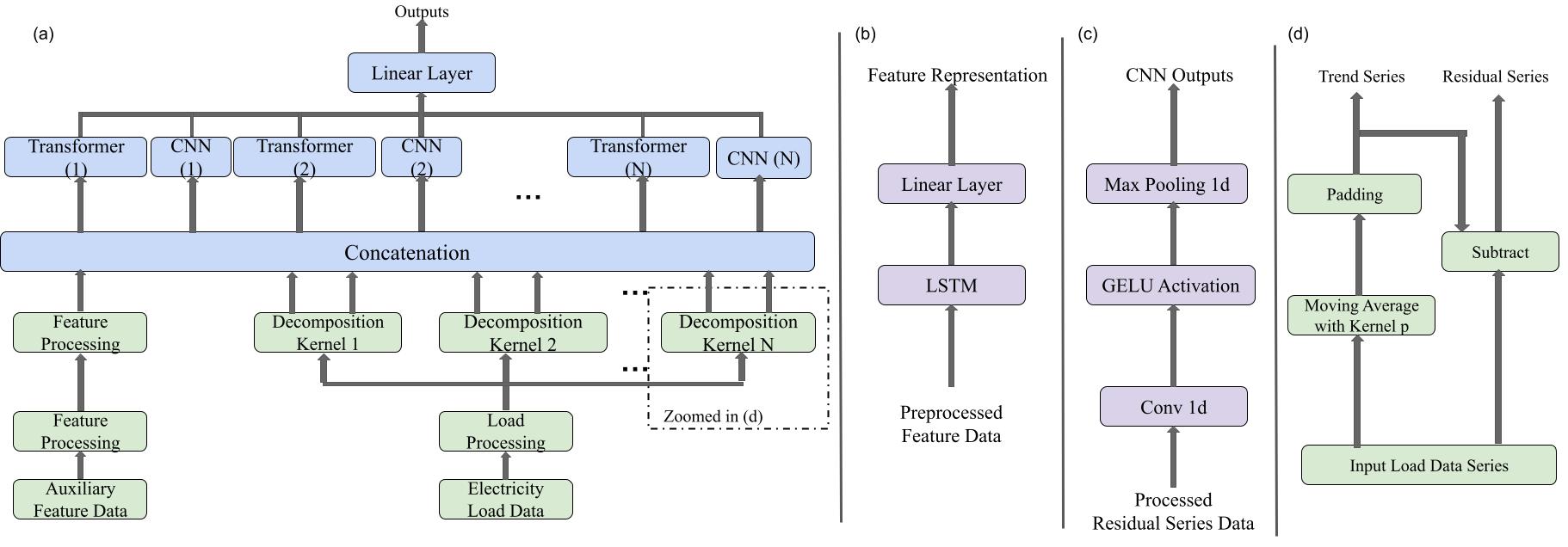}
    \caption{Overall framework of the proposed model. (a) illustrates the overall structure of the proposed method, where the temporal data decomposition and auxiliary feature representation are carried out separately. In (b), it illustrate the auxiliary feature representation. For each representation, the model adopt an LSTM and then a linear layer is utilized to learn their combinations. In (c), we show the utilized CNN structure in the proposed model. In (d), a diagram of the temporal decomposition is illustrated. }
    \label{framework}
\end{figure*}

%\textcolor{red}{This is a simple organization paper. No need to do this introduction.}The rest of the paper is organized as follows: Section \ref{task} demonstrates the task and the overall structure of our proposed model. In Section \ref{model} we will a detailed description of the model mechanism. In Section \ref{exp}, we show the performance of our proposed model and compare it with other baselines. We also carry out ablation and generalization studies to verify the effectiveness of our method. In Section \ref{conclusion}, we summarize our method. 

\section{Task Description and Model Framework}
\label{model}

\subsection{Datasets And Forecasting Tasks}
\label{task}
Initially, the task is to forecast the future $T$ steps utilizing the previous $P$ steps load feature records. The dataset $\mathcal{D} \in \mathbb{R}^{M \times (1+K)}$ is collected from Belgium load data starting from 1st January 2018 to 31st December 2019. For the test experiment, two datasets is also included: $\text{Test}_1$ is collected from Belgium load data starting from 1st January 2020 to 31st December 2021; $\text{Test}_2$ is collected from New South Wales from Australia load data starting from 1st January 2020 to 31st December 2021. with a time resolution of 1 hour. The dataset consists of two parts, where the target part is the electricity load data $\mathcal{L} \in \mathbb{R}^{M\times 1}$ and the auxiliary feature set is $\mathcal{F} \in \mathbb{R}^{M\times K}$, where $K$ denotes the total number of auxiliary features. In the applied dataset, $K=6$ features are included, such as temperature, calendar, and humidity. \textcolor{blue}{The number of data samples is $M=17520$}. It is noticeable that the dataset is normalized, where it scaled up by its mean and variance. In the numerical study, the experiment with the actual data is also included and the reason for scaling is also discussed empirically. Afterward, positional and token embedding are utilized which can provide a rough sense of long-dependencies for the model \cite{vaswani2017attention}. For training, validating, and testing, the dataset with the length of $P$ are sampled as the original input and we denote each sampled load and feature series as $L_i \in \mathbb{R}^{P \times 1}$ and $F_{i} \in \mathbb{R}^{P \times K}$, where $i=1, 2, ..., M-P+1$ denotes the $i-th$ sampled load and feature series. 

The dataset is split into train, validation, and test sets with a ratio of 7:2:1. Thus, the training set can be represented as $\mathcal{D}^{train} \in \mathbb{R}^{M^{tr}\times (1+K)}$ (Similar for $\mathcal{D}^{val}$ and $\mathcal{D}^{test}$, and thereafter, $\{\cdot\}^{tr}$, $\{\cdot\}^{val}$, $\{\cdot\}^{test}$ are also used to splitting the train, validation, and test set for the  $\mathcal{L}$ and $\mathcal{F}$. \textcolor{blue}{The precise numbers of the data samples for train, validation, and test set are $M^{tr}=12264$, $M^{val}=3504$, and $M^{test}=1752$ respectively}).

\subsection{Interpretability and Model Overview}
In this paper, a post-hoc global interpretability method is proposed without any dataset dependencies. The proposed method illustrates both temporal and feature interpretability, and their contributions to the model outputs are visualized by heatmaps. Noticeably, the temporal and the feature interpretabilities are obtained from the feature engineering, which also benefits the forecasting accuracy. 

Temporal interpretability is derived from the decomposition operations. It represents how the past load sequence affects the future load. To this end, we propose a multi-scale temporal decomposition to break down the load series into various components, which will be illustrated in Section \ref{decomposition}. 

As for the feature engineering and interpretability, to tackle the temporal dynamics in these feature series, a one-layer LSTM is applied to learn their representations and a linear layer to learn their combinations for further representations\cite{li2020exploring}. Thus, before training task-relevant features are enhanced, and task-irrelevant features are partially suppressed~\cite{hermann2020shapes}. See details in Section \ref{feature engineering}. 

After decomposition and feature engineering, the detailed forecasting strategy is illustrated by incorporating a group of Transformers and CNNs modules in Section \ref{model forecast}. We can visualize the significance of the temporal and auxiliary features to the output prediction in the format of heatmaps \cite{olah2017feature}. In that case, it can be clearly demonstrate how important the global features are for future prediction.

\section{Multi-Scale Temporal Decomposition}
\label{decomposition}
In this section, a detailed demonstration of the key components of the framework will be explained, which will include the decomposition kernels, auxiliary feature engineering, and how the model makes forecasts. Multi-scale temporal decomposition enables us to have various temporal components with different time scales. %Empirically, we will also illustrate that decomposition kernels are beneficial for forecasting accuracy in Section \ref{exp}.
Besides, an auxiliary feature representation is also carried out, and the processed representation is concatenated with the temporal feature data to form the  input for final layers of the proposed model as illustrated in Fig. \ref{framework}. %Various neural Transformers and CNNs are utilized to the input and the last linear layer will learn a combination of the outputs. 

\subsection{Decomposition Kernel Settings}
\label{decomposition}
Given a long sequence of time series, we are interested in first analyzing the temporal patterns pertaining to the data. The proposed method can take advantage of the temporal decomposition in GAM and the strategy to train multiple NNs to learn different features. In the proposed method, unlike most previous methods, instead of applying the original load data for the models to learn, multi-scale temporal decomposition kernels are utilized to process the load data~\cite{zeng2023transformers}. Inspired by NAM, the proposed method also concentrate on multi-timescale temporal feature interpretability by training multiple NNs. After the decomposition, multi-scale trend and seasonal features can be obtained. Specifically, different sizes of decomposition kernels are applied. Each time a decomposition kernel is applied, it can get two components: the trend part and the residual seasonal part. In addition, applying models directly to learn the raw time series data requires a complex structure to handle both the trend and residual patterns, like \cite{zhou2022fedformer}. Many time series exhibit clear trend patterns and decomposition can separate the components, providing models with a clearer view of underlying patterns \cite{wu2021autoformer}. Decomposing a time series can lead to a clearer understanding of its temporal components. By modeling each component separately, models can provide insights into the different aspects of the data. %To obtain the trend part, we carry out moving average operations to the original load, and the residual series was obtained by using the trend part to subtract the original load.

\begin{figure}[tp]
    \centering
    \includegraphics[scale=0.267]{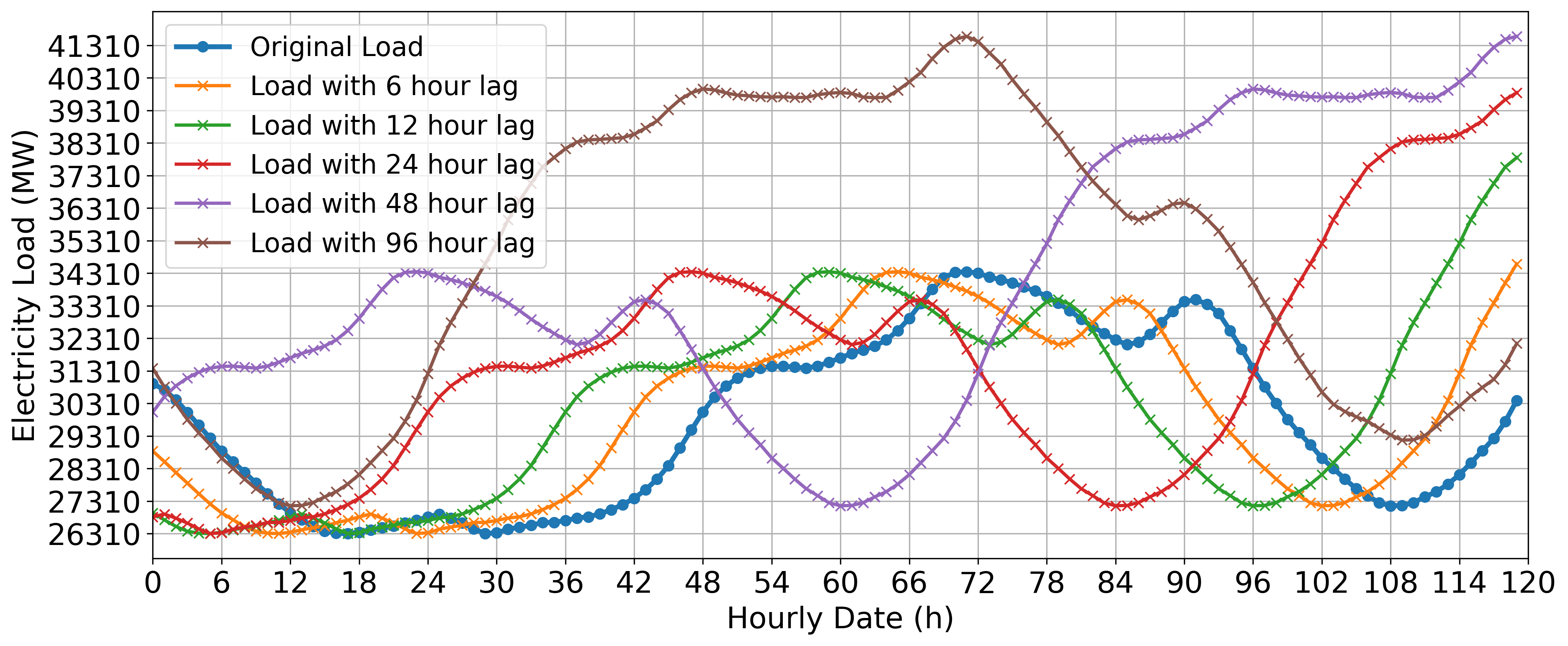}
    \caption{The current load series at a specific time point and its lagged series. As can be derived from the cosine similarity heatmap, these lagged series illustrate high similarities with the current series.}
    \label{cosine}
\end{figure}

The decomposition kernel is one of the key components that boost the model forecasting accuracy and also illustrates the temporal interpretability. To figure out the configurations of the decomposition kernels, initially, cosine similarity is applied to roughly measure the temporal patterns of the load data. During training, the sampling length from the original load data is $P$, thus the model aim to find out the temporal patterns associated with the $P$ sampling length. 

\begin{figure}[tp]
    \centering
    \includegraphics[width=1.0\linewidth]{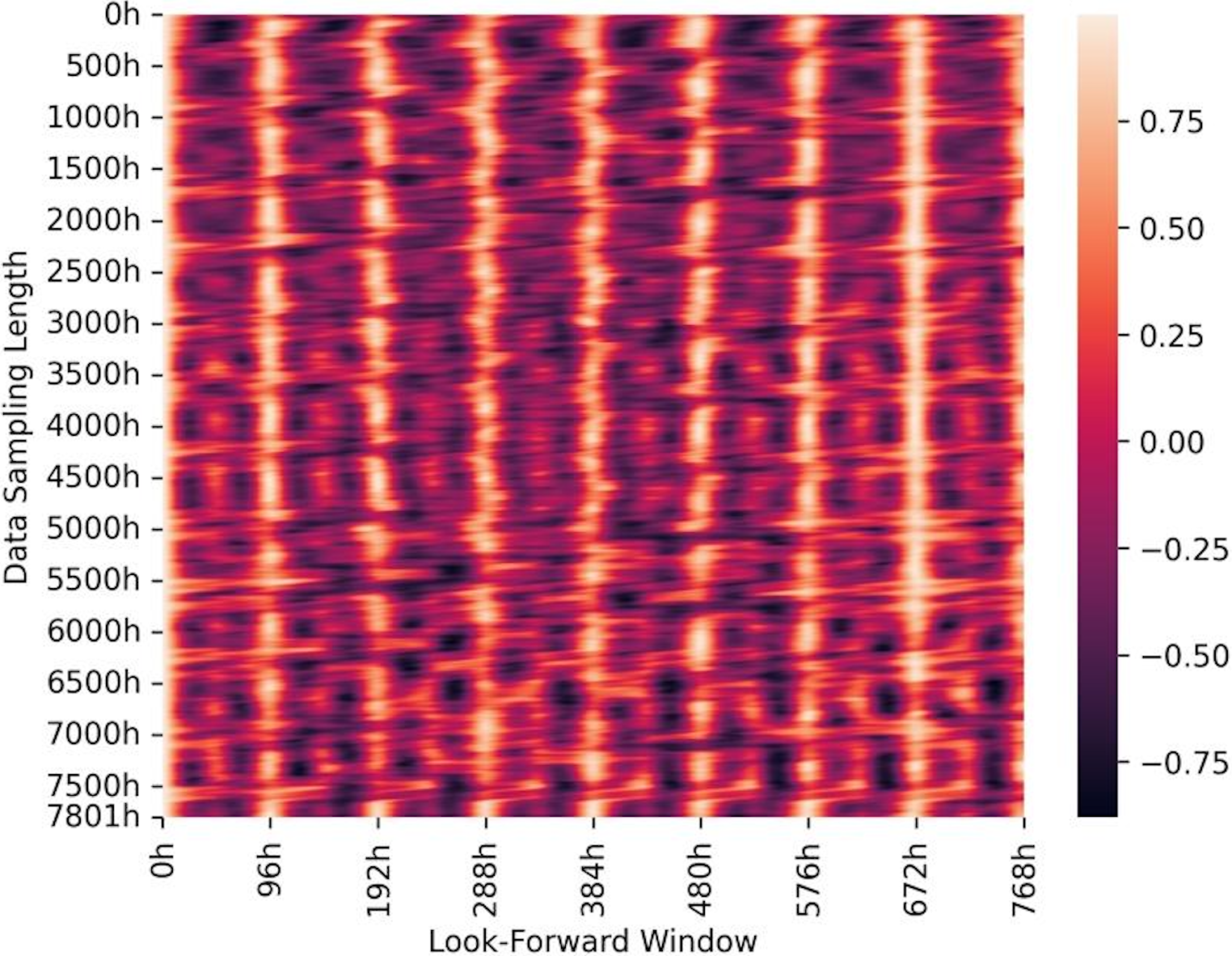}
    \caption{Temporal dependencies calculated in \eqref{equ:similarity} for the Belgium load data.}
    \label{Heat1}
\end{figure}

The sampling principle is that for each time step of the total $M^{tr}$ time steps, it sample a series with the length of $P$ and it move forward for one step until the original load series is all sampled. As previously mentioned, the sampled load data $L_i$ can be derived from $\mathcal{L}$. Then $L_i$ is used to calculate the similarities with its future samples. The look forward window for the calculation is denoted as $W$, and for the convenience of visualization, we set $W=768$. Thus, a total of $(M^{tr}-P-W+1)$ samples can be obtained via the following cosine similarity calculation in Eq. \ref{equ:similarity} as
\begin{equation}
\label{equ:similarity}
    Sim(L_i, L_j) = \frac{L_i \cdot L_j}{\|L_i\|_2 \times \|L_j\|_2},
\end{equation}
where $L_i$ is the current sampled series at time point $i$, $i=1, 2, ..., (M^{tr}-P-W+1)$, $j=i+1, ..., i+W$, and $||\cdot||_2$ is the Euclidean norm. If the absolute value of $Sim(L_i, L_j)$ is close to 1, then the correlation between the two series is strong, which indicates a similar pattern. To get a rough sense of the temporal patterns, in Fig. \ref{cosine}, a load series at a specific time and its lagged series is visualized. Furthermore, to be more precise about the temporal patterns, by quantitatively analyzing Fig. \ref{Heat1}, we can obtain more significant similarity values and trend patterns at 6, 12, 24, 48, and 96 hours. Based on such observations, $N$ decomposition kernels are selected for the later decomposing operations. This helps reveal the periodicity features associated with the sampling length selection. 

The decomposition method is based on the idea of moving average and the moving average kernel size is the moving window size \cite{zeng2023transformers}. The kernel size of the decomposition depends on the periodicity patterns of $\mathcal{L}$ and the sampling length $P$, which is derived from the previous cosine similarity operations carried out for sampling length $P$. In order to provide multiple temporal patterns for better interpretability and enable the model to have both short-term and long-term features to learn, $N$ decomposition kernels are selected and the kernel size is denoted as $K_m, m=1, 2, ..., N$. The decomposed trend series can be expressed as:
\begin{equation}
    \mathcal{T}^ (n)=\begin{cases} 
  \mathcal{L}^ (1), &  n <=\frac{K_m-1}{2}  \\
 \frac{1}{K_m}\sum_{p=n-\frac{K_m-1}{2}}^{{n+\frac{K_m-1}{2}}}\mathcal{L}_p,  &  \frac{K_m-1}{2}<n<M^{tr}-\frac{K_m-3}{2} \\
  \mathcal{L}(M^{tr}), &  n >= M^{train}-\frac{K_m-3}{2}
\end{cases}
\end{equation}
where $\mathcal{L}(1)$ and $\mathcal{L}(M)$ are the load values at the first and last time step of $\mathcal{L}$, $n=1, 2, ..., M^{tr}$. $\mathcal{L}(1) \:\text{and}\: \mathcal{L}(M^{tr})$ are utilized for padding the obtained trending data $\mathcal{T} \in \mathbb{R}^{M^{tr}\times 1}$. Naturally, if the moving average operation is applied to a series with the kernel size of $K_m$, the length of the series will be shortened for $(K_m-1)$. To align the series and maintain the original length after decomposition (moving average), we use $\mathcal{L}(1)$ and $\mathcal{L}(M)$ to pad at the head and the rear of the decomposed series, with the length of $\frac{K_m-1}{2}$ respectively\cite{zeng2023transformers}. To get the residual seasonal series, the original series is used to subtract the obtained trend series which can be expressed as:
\begin{equation}
    \mathcal{R}(n)=\mathcal{L}(n)-\mathcal{T}(n),
\end{equation}
where $\mathcal{R}(n) \in \mathbb{R}^{M^{tr} \times 1}$ is the overall residual seasonal data, $n=1, 2, ..., M^{tr}$. Each time a decomposition operation is carried out, it can obtain a trend and a residual series whose shapes are both $[M^{tr}, 1]$. When $N$ different decomposition kernels are utilized, $N$ trend series and $N$ residual series can be obtained, where we denote them as $\mathcal{T}_p$ and $\mathcal{R}_p$, $p=1, 2, ..., N$. As for the input of the model, we sample these series with the length $P$ and denote them as $T_{pr}$ and $R_{pr}$, $r=1, 2, ..., M^{tr}$. 

\subsection{Auxiliary Feature Process}
\label{feature engineering}
After decomposing and obtaining the load data required for the model training, the model also need to do auxiliary feature engineering to learn representations. The learned representations will be concatenated with the decomposed load data to form the data input for the model.

In this subsection, we will draw attention to the interpretable auxiliary feature engineering. Auxiliary features are crucial for promoting forecasting accuracy for they can provide additional context that a single time-series forecasting model cannot capture. For feature series $\mathcal{F}^{tr}$, the feature with length $P$ is sampled to get the input series $F_{r}$, $r=1, 2, ..., M^{tr}$. To tackle such temporal-dependent sequences, $K$ one-layer LSTM is utilized for each feature $F_{qr}$ to learn their corresponding representations respectively as below:
\begin{equation}
    F'_{qr}=\text{LSTM}_q(F_{qr}),
\end{equation}
where $F'_{qr} \in \mathbb{R}^{P \times 1}$ is the $q$-th $LSTM_q$-processed feature for the $r$-th sample, $q=1, 2, ..., K$. 

Afterward, the model also utilize a linear layer to learn a linear combination of each representation and thus it can obtain the significance of the different feature representations with the shape of $[M^{tr}, 1]$. The auxiliary feature representation of the $r$-th sample can be expressed as:
\begin{equation}
    F'_r=\sum_{q=1}^N\gamma_q \times F'_{qr};
\end{equation}
where $\gamma_q$ is the significance of the $q$-th features, which is trainable, and $F'_r \in \mathbb{R}^{P \times 1}$ is the feature representation of the $r$-th sample. By reading $\gamma_q$, the significance of the auxiliary features can be derived. 

\subsection{Model Forecasting}
\label{model forecast}
The various decomposition kernels enable us to have trend and residual features of different timescales. Since the trend part represents the underlying direction in which data is moving, typically over a long period, and can therefore be seen as a low-frequency feature, while high-frequency features would correspond to more rapid, often periodic fluctuations, such as daily or seasonal variations\cite{daubechies1990wavelet}. 

The proposed model is inspired by~\cite{park2022vision}, which mathematically and empirically prove that neural Transformer and CNN are specialized in learning low-frequency and high-frequency features respectively. Thus, the model can take advantage of the neural Transformer to learn the low-frequency features (trend part) and CNN to learn the high-frequency features (residual part). Neural Transformer is specialized in capturing long-term dependencies with the self-attention mechanism. On the other hand, CNNs designed with their local feature extraction capability, are adept at capturing short-term and high-frequency patterns in the data~\cite{park2022vision}. Based on this intuition, the model can utilize various Transformers and CNNs to attend to different temporal features after multi-scale time series decomposition. Afterward, a linear layer is alos applied to learn the significance of each temporal component. \textcolor{blue}{Specifically, the linear combination of the temporal components learned by different Transformers and CNNs will provide the direct interpretability of the temporal features. Once trained,  $\hat{y}^r$ learns the linear combination from the output of each module $TF_p$ and $CN_p$.} The empirical study demonstrates that the proposed decomposition method not only can provide interpretable features for the proposed model to learn, but also boost the prediction accuracy. 

After auxiliary feature representation and temporal decomposition operation, we can construct transformed forecasting samples by concatenating the learned feature representations and the temporal feature series as follows:
\begin{align}
    T'_{pr}(t) & =\text{Concat}(T_{pr}(t), F'_r(t)), \\
    R'_{pr}(t) & =\text{Concat}(R_{pr}(t), F'_r(t)),
\end{align}
where $T'_{pr}$ and $R'_{pr}$ are the updated $r$-th sample input for the model, $p=1, 2, ..., N$, and $t$ is the time step ranging from $1$ to $P$. As previously mentioned, $T_{pr}(t)$ and $R_{pr}(t)$ have the same shape of $[P, 1]$ and after concatenation, the shape of the new input $T'_{pr}(t)$ and $R'_{pr}(t)$ is $[P, 2]$. To tackle the trend and residual part, $N$ Transformers \cite{vaswani2017attention} (denoted as $TF_p, p=1, ..., N$) and $N$ CNNs (denote as $CN_p, p=1, ..., N$) are applied to learn the trend part $TF_p(T'_{pr})$ and residual seasonal details $CN_p(R'_{pr})$ respectively~\cite{park2022vision}. 

The structure of the applied CNN is shown in Fig. \ref{framework}. After obtaining the $2N$ outputs from $TF_p$ and $CN_p$, a linear layer is further utilized to learn their significance scores, which will provide the temporal pattern interpretability. The outputs of the model can be expressed as: 
\begin{equation}
    \hat{y}_r=\sum_{p=1}^{N}\alpha_pTF_p(T'_{pr})+\sum_{p=1}^N\beta_pCN_p(R'_{pr});
\end{equation}
where $\alpha_{p}$ and $\beta_{p}$ are the trainable significance scores of the outputs from $TF_p$ and $CN_p$, and $\hat{y}_r \in \mathbb{R}^{T \times 1}$ is the forecasting output vector. \textcolor{blue}{Specifically, $\hat{y}_r$ learns the linear combination from the output of the $TF_p$ and $CN_p$.} To update the parameters in the model, Mean Squared Error (MSE) is used as the loss function, which can be expressed as: 
\begin{equation}
\label{equ:MSE}
    MSE = \frac{1}{B} \sum_{r=1}^B\|\hat{y}_r-y_r\|_2^2;
\end{equation}
where $B$ is the batch size during training; $\hat{y}_r$ and $y_r$ are the forecasting values and the real values respectively. To get the ground truth value for the loss function, $y_r=L_r$, which is sampled from $\mathcal{L}$ with the length $T$, $r=1, 2, ..., M^{tr}$. 

\begin{figure}
    \centering
    \includegraphics[width=0.45\textwidth]{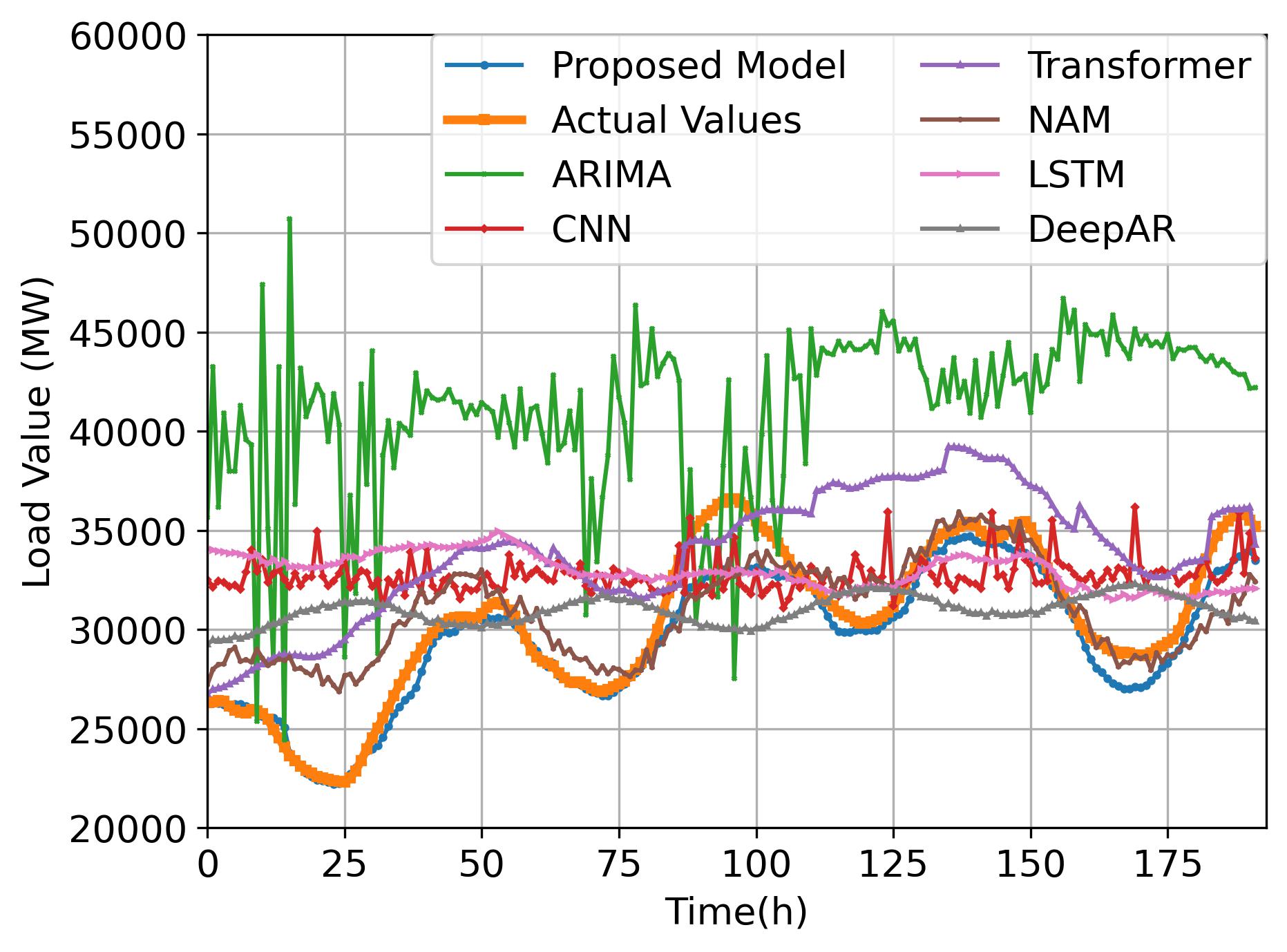}
    \caption{Hourly load forecasting visualization}
    \label{forecast visualization}
\end{figure}

\textcolor{blue}{\subsection{Complexity Analysis}}
\textcolor{blue}{Some previous interpretable SOTA method usually lead to higher time complexity \cite{sundararajan2020many, garreau2020explaining}. Thus, it is necessary to analyze the time complexity for the proposed interpretable method and $O$ complexity will be discussed in this subsection (given the dataset $\mathcal{D} \in \mathbb{R}^{M \times (1+K)}$). }

\textcolor{blue}{For temporal decomposition, $N$ different decomposition kernels are utilized with the size of $K_m, m=1, 2, ..., N$. The padding length for the series is $K_m-1$, so the total padding cost is $O(N*(K_m-1))$. As for the moving average operation, at each time step, the cost is $O(K_m)$, and the moving stride is 1. Thus, the cost for moving average is $O(\sum_{M=1}^N M*K_m)$. 
For the auxiliary feature processing, $K$ LSTMs are trained to tackle the $K$ total features, where the LSTM has one layer. Thus, all $K$ LSTMS cost $O(K*M*h*P)$~\cite{karim2017lstm}. }

\textcolor{blue}{After obtaining the processed feature representation and the decomposed temporal components, the concatenation operation is utilized as shown in Fig. \ref{framework}, and the total cost is $O(N*M)$. For each Transformer with $C$ layers, its complexity is $O(C*M*d^2)$. All $N$ CNNs have computations of $O(2*N*G*M*F)$. After obtaining the results from all $2N$ Transformers and CNNs, a linear combination is also learned to get the final result, and the cost is $O(N*M)$. Notice $K_m, K, N, C, G, F <<M$. Thus, the total cost can be approximately calculated as $O(H*(M*(h^2+d^2)+d*M^2))$, where $H$ is a constant. Given that $M >> h, d$, the final cost can be estimated as $O(H*d*M^2)$, which is quadratic with respect to the length of the dataset.}

\section{Result Evaluations and Discussions}
\label{exp}

\subsection{Baseline Experiments}
To measure the forecasting performance of the model, apart from MSE, Mean Absolute Error (MAE), Root Mean Squared Error (RMSE), and Mean Average Percentage Error (MAPE) are also adopt as the metrics:
\begin{align}
    MAE & =\frac{1}{M^{test}} \sum_{i=1}^{M^{test}}\|\hat{y}_i-y_i\|_1;  \label{MAE} \\
    RMSE &= \sqrt{\frac{1}{M^{test}} \sum_{i=1}^{M^{test}}\|\hat{y}_i-y_i\|_2^2} \label{rmse}; \\
    MAPE &=\frac{100\%}{M^{test}} \sum_{i=1}^{M^{test}}\|\frac{\hat{y}_i-y_i}{y_i}\|_1 .\label{mape}
\end{align}

To evaluate the performance of the proposed method, other baseline models are also included, like the Transformer and CNN, which are applied in the proposed method, and also the most common autoregressive method, ARIMA. NAM is also included as the baseline model and we will make comparisons about both interpretability. Besides, classical sequence forecasting models LSTM and the latest DeepAR are also utilized for baselines. In Table \ref{tab: 1}, the error metrics is given out and in Fig. \ref{forecast visualization}, the forecasting result is visualized. We also carry out the experiment where the data is not standardized and the forecasting accuracy is displayed in Table \ref{tab: 2}.

\textcolor{blue}{\subsection{Results discussion}}
\textcolor{blue}{As shown in Table \ref{tab: 1} and Table \ref{tab: 2}, the proposed method demonstrates the best forecasting errors when trained on both standardized and actual datasets (in MW). The average MSE, MAE, RMSE and MAPE of the proposed method using the standardized training data are 0.52, 0.57, 0.72 and 2.16\% respectively. The additive model NAM performs the second best in terms of the forecasting accuracy. Another interpretable model DeepAR performs similarly with LSTM. The interpretability of DeepAR is based on probabilistic forecasting, which is different with the proposed method and NAM. Classic autoregressive method ARIMA has the highest forecasting errors, which is caused by its incapability of handling complicated non-linearity patterns of load data. }

\textcolor{blue}{Numerically, as for MSE, MAE, and RMSE, suppose the forecasting values trained with standardized data is the standardization of the forecasting results trained with real data, based on Eq. \eqref{equ:MSE} and \eqref{MAE}, it can be analyzed as:}
\begin{align}
MSE_{scaled} &= \frac{1}{M^{test}} \sum_{i=1}^{M^{test}} \| \frac{\hat{y}_{i} - \mu^{test}}{\sigma^{test}} - \frac{y_{i} - \mu^{test}}{\sigma^{test}} \|_2^2 \nonumber \\
&= \frac{1}{(\sigma^{test})^2} \cdot \frac{1}{M^{test}} \sum_{i=1}^{M^{test}} \| \hat{y}_{i} - y_{i}\|_2^2 \nonumber \\
&= \frac{MSE}{(\sigma^{test})^2}; \\
RMSE_{scaled} &= \sqrt{MSE_{scaled}}  
= \frac{RMSE}{\sigma^{test}};
\end{align}
\begin{align}
    MAE_{scaled} &= \frac{1}{M^{test}} \sum_{i=1}^{M^{test}} \| \frac{\hat{y}_{i} - \mu^{test}}{\sigma^{test}} - \frac{y_{i} - \mu^{test}}{\sigma^{test}} \|_1 \nonumber \\
&= \frac{1}{M^{test}} \sum_{i=1}^{M^{test}} \| \frac{\hat{y}_{i} - y_{i}}{\sigma^{test}} \|_1 \nonumber 
= \frac{MAE}{\sigma^{test}};
\end{align}
\textcolor{blue}{where $\mu^{test}$ and $\sigma^{test}$ are the mean and standard deviation of the actual load data of the test set. Under this assumption, $MSE_{scaled}$ should scale quadratically with $MSE$. $MAE_{scaled}$ and $RMSE_{scaled}$ should scale linearly with $MAE$ and $RMSE$. However, it can be seen that the actual results $MSE_a$, $MAE_a$, and $RMSE_a$ are much larger than the theoretical $MSE$, $MAE$, and $RMSE$. Thus, it is proved that applying standardization can reduce the absolute test error. }

\textcolor{blue}{As for MAPE value, based on Eq. \ref{mape}, it can be shown that $MAPE_{scaled}$ is not scaled with linearly with the theoretical $MAPE$:}
\begin{align}
    MAPE_{scaled} = \frac{100\%}{n} \sum_{i=1}^{n} \| \frac{\frac{\hat{y}_{i} - \mu^{test}}{\sigma^{test}} - \frac{y_{i} - \mu^{test}}{\sigma^{test}}}{\frac{y_{i} - \mu^{test}}{\sigma^{test}}} \|_1.
\end{align}

\textcolor{blue}{However, the percentage errors could appear relatively small because the absolute errors (the numerator in the MAPE formula) are divided by very large actual values (the denominator in the MAPE formula). This can make the MAPE value seem small even if the absolute errors are significant in real terms. Thus, MAPE metric may not reflect the absolute forecasting errors well and can be influenced by the training data. }

\begin{figure}
    \centering
    \includegraphics[width=0.4\textwidth]{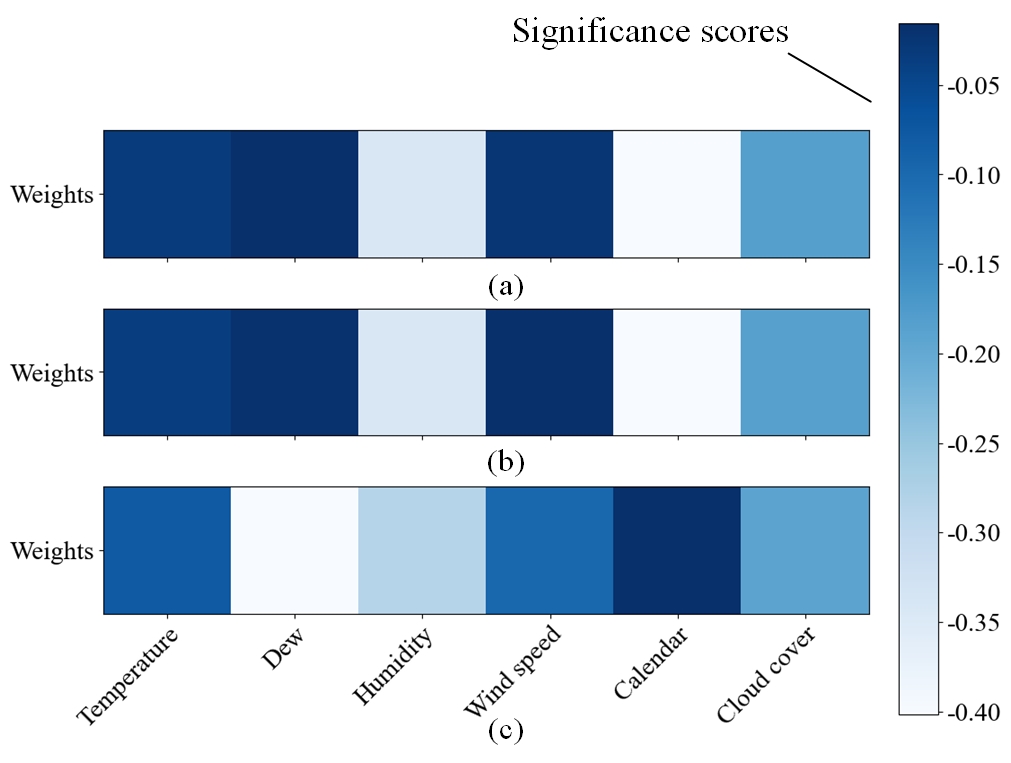}
    \caption{(a) The auxiliary feature significance scores of the Belgium dataset. (b) and (c) The feature significance score of the test datasets $\text{Test}_1$ and $\text{Test}_2$.}
    \label{Heat2}
\end{figure}

\begin{figure}
    \centering
    \includegraphics[width=0.4\textwidth]{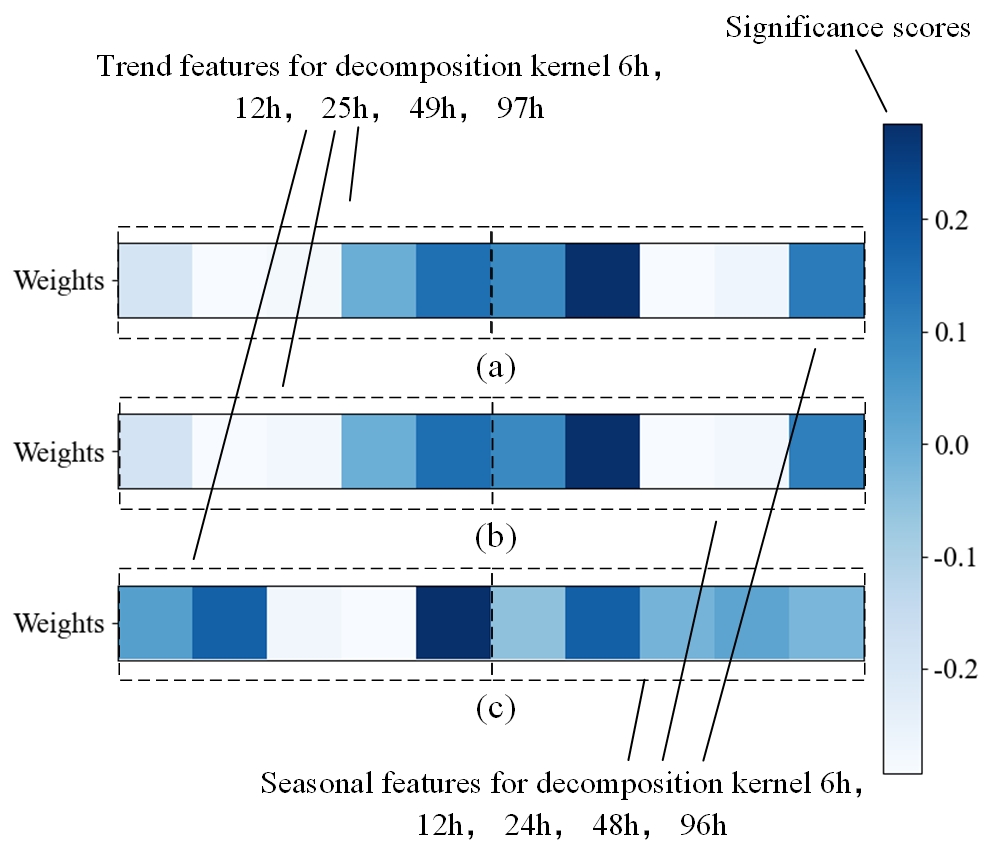}
    \caption{(a) illustrates the significance scores of the temporal features of the Belgium dataset. (b) and (c) display the significance score of the temporal features of the test datasets $\text{Test}_1$ and $\text{Test}_2$.}
    \label{Heat3}
\end{figure}

% \begin{figure}
%     \centering
%     \includegraphics[width=0.75\linewidth]{nam.png}
%     \caption{Feature weights for NAM. Settings same as Fig. \ref{Heat2}.}
%     \label{nam}
% \end{figure}

As for feature and temporal interpretabilities, the significance scores of the auxiliary and temporal features is visualized in Fig. \ref{Heat2} and Fig. \ref{Heat3} through heatmaps. The color encoding enable us to analyze the relative significance of different auxiliary and temporal features. From these two figures, factors like humidity and calendar features are the most significant two features related to the model outputs. 

\textcolor{blue}{Table \ref{tab:3} lists the significance scores of NAM and the proposed proposed model. In NAM, temporal decomposition is not utilized and the load data is treated as an input feature. Since load data is the forecasting target, the load is the most important feature for NAM.} Similarly, apart from the load feature, calendar, and humidity features are significant features for forecasting. Compared with the proposed model, though both models can provide global interpretability, NAM fails to provide more detailed temporal features. Besides, the proposed model can outperform NAM in forecasting accuracy. 

\begin{table}[tp]
    \centering
    \caption{Forecasting accuracy of the standardized Belgium load data}
    \label{tab: 1}  
    \scalebox{0.88}{
    \begin{tabular}{lllll}
        \hline\noalign{\smallskip}    
        Model & MSE-scaled & MAE-scaled & RMSE-scaled & MAPE-scaled (\%)  \\
        \noalign{\smallskip}\hline\noalign{\smallskip}
        Proposed & $\textbf{0.52}(\pm 0.20)$ & $\textbf{0.57}(\pm 0.13)$ & $\textbf{0.72}(\pm 0.15)$ & $\textbf{2.16}(\pm 0.36)$  \\
        NAM & $0.63(\pm 0.11)$ & $0.64(\pm 0.09)$ & $0.80(\pm 0.07)$ & $3.89(\pm 0.23)$ \\
        LSTM & $0.78(\pm 0.09)$ & $0.68(\pm 0.08)$ & $0.87(\pm 0.03)$ & $3.97(\pm 0.61)$ \\
        DeepAR & $0.80(\pm 0.15)$ & $0.71(\pm 0.09)$ & $0.90(\pm 0.05)$ & $4.01(\pm 1.07)$ \\
        Transformer & $0.95(\pm 0.25)$ & $0.71(\pm 0.09)$ & $0.97(\pm 0.13)$ & $4.10(\pm 1.15)$ \\
        
        CNN & $1.04(\pm 0.19)$ & $0.82(\pm 0.08)$ & $1.02(\pm 0.09)$ & $7.47(\pm 0.71)$ \\
        ARIMA & $3.58(\pm 2.54)$ & $1.64( \pm 1.81)$ & $1.89(\pm 1.62)$ & $11.84 (\pm 1.73)$ \\
                \noalign{\smallskip}\hline
    \end{tabular}}
\end{table}

\begin{table}[tp]
	\centering
	\caption{Forecasting accuracy of the actual Belgium load data}
	\label{tab: 2}  
	\scalebox{0.9}{
	\begin{tabular}{lllll}
		\hline\noalign{\smallskip}	
		Model & MSE-a ($\times 10^8$) & MAE-a ($\times 10^3$) & RMSE-a ($\times 10^3$) & MAPE-a (\%)  \\
		\noalign{\smallskip}\hline\noalign{\smallskip}
	    Proposed   & $\textbf{1.81}(\pm 0.11)$ & $\textbf{3.48}(\pm 0.06)$ & $\textbf{4.25}(\pm 0.26)$ & $\textbf{0.10}(\pm 0.01)$  \\
        NAM & $1.83(\pm 0.03)$ & $3.63(\pm 0.12)$ & $4.36(\pm 0.17)$ & $0.12(\pm 0.01)$ \\
        LSTM & $2.11(\pm 0.09)$ & $4.31(\pm 0.07)$ & $4.42(\pm 0.01)$ & $0.98(\pm 0.01)$ \\
        DeepAR &$2.11(\pm 0.12)$ &$4.30(\pm 0.09)$ &$4.43(\pm 0.01)$ &$0.99(\pm 0.01)$ \\
        
		Transformer & $2.20(\pm 0.02)$ & $4.33(\pm 0.04)$ & $4.43(\pm 0.01)$ & $1.00(\pm 0.02)$ \\
        
		CNN & $2.82(\pm 0.04)$ & $4.50(\pm 0.07)$ & $4.72(\pm 0.09)$ & $1.06(\pm 0.03)$ \\
        ARIMA & $5.38(\pm 1.24)$ & $4.39(\pm 1.81)$ & $8.52(\pm 1.62)$ & $1.76 (\pm 0.35)$ \\
				\noalign{\smallskip}\hline
	\end{tabular}}
\end{table}

As for the temporal features, the trend and residual details of 6, 12, and 24 hours are critical to the forecasting results. For CNN, utilizing CNN solely to forecast lacks direct interpretability compared with the proposed method. The convolution operation filters the input data into a set of representations. However, understanding the significance of each filter or specific input features based on the convolutions is non-trivial. As for ARIMA, it can provide good interpretability on the temporal and auxiliary features, while they generally perform worse on forecasting compared with other baseline models. \textcolor{blue}{The proposed method also achieves superiority over Transformer. On the accuracy perspective, we have empirically illustrates that the temporal decomposition can boost the forecasting accuracy. As for the interpretability perspective, the adopted auxiliary feature engineering operation and temporal decomposition ensure the feature and temporal interpretability, while the self attention mechanism of the plain Transformer is hard to explain. }

\begin{table}[tp]%调节图片位置，h：浮动；t：顶部；b:底部；p：当前位置
	\centering
	\caption{Feature significance scores of the proposed model and NAM}
	\label{tab:3}  
	\scalebox{1.0}{
	\begin{tabular}{lll}%表格中的数据居中，c的个数为表格的列数
		\hline\noalign{\smallskip}	
		Model & Proposed & NAM  \\
		\noalign{\smallskip}\hline\noalign{\smallskip}
	    Temperature   & $-0.0328$ & $-0.0369$ \\
        Dew & $-0.0155$ & $0.1955$\\
        Humidity & $-0.3431$ & $-0.3313$ \\
        Wind speed & $-0.0252$ & $0.2629$ \\
        Calendar & $-0.4016$ & $-0.2854$ \\
        Cloud cover & $-0.1839$ & $-0.2976$\\
        Load & Decomposed & $0.4257$\\
        \noalign{\smallskip}\hline
	\end{tabular}}
\end{table}

\subsection{Analysis of Model Parameters and Interpretability}
%As electricity load forecasting is a highly data-driven task, the framework of our proposed method can also be flexible. In other words, 
In the model design, the number of the decomposition kernels $N$ can be adjusted. In Experiment 2-5 listed in Table \ref{tab: 4}, we empirically show the reason for selecting $N=5$ decomposition kernels, where the selection depends on forecasting accuracy. Due to the limitations of computation resources, experiments with $N\geq 6$ decomposition kernels will not be carried out. As there always involve a trade-off between accuracy and interpretability, in Experiment 2, while the $N=4$ kernel exhibits a slightly higher accuracy compared to the proposed, this improvement comes at the cost of reduced interpretability due to the loss of two temporal features. Compared to Experiments 3, 4, and 5, where 3, 2, and 1 decomposition kernels are used respectively, the proposed model has achieved both better interpretability and accuracy.  

\begin{table}[tp]%调节图片位置，h：浮动；t：顶部；b:底部；p：当前位置
	\centering
	\caption{Study on Decomposition Kernels}
	\label{tab: 4}  
	\scalebox{0.76}{
	\begin{tabular}{lllll}%表格中的数据居中，c的个数为表格的列数
		\hline\noalign{\smallskip}	
		Model & MSE & MAE & RMSE & MAPE(\%) \\
		\noalign{\smallskip}\hline\noalign{\smallskip}
	    Experiment 1 (Proposed)  &  $0.52(\pm 0.20)$ & $0.57(\pm 0.13)$ & $0.72(\pm 0.15)$ & $2.16(\pm 0.36)$  \\
		Experiment 2 (N=4) & $0.49(\pm 0.28)$ & $0.53(\pm 0.14)$ & $0.70(\pm 0.19)$ & $2.14(\pm 0.68)$\\
		Experiment 3 (N=3) & $0.53(\pm 0.14)$ & $0.56(\pm 0.11)$  & $0.72(\pm 0.17)$ &$2.88(\pm 0.89)$ \\
        Experiment 4 (N=2) & $0.58(\pm 0.18)$ & $0.59(\pm 0.08)$ & $0.76(\pm 0.11)$ 
 & $3.15(\pm 0.76)$\\
        Experiment 5 (N=1) & $0.57(\pm 0.19)$ & $0.57(\pm 0.09)$ & $0.75(\pm 0.14)$ & $2.91(\pm 0.94)$ \\
        Experiment 6 (5 TF) & $0.94(\pm 0.11)$ & $0.78(\pm 0.05)$ & $0.97(\pm 0.06)$ 
 & $3.52(\pm 0.71)$\\
        Experiment 7 (5 CN) & $1.05(\pm 0.14)$ & $0.83(\pm 0.07)$ & $1.03(\pm 0.08)$ & $3.56(\pm 0.53)$ \\
        Experiment 8 (5 TF and 5 CN) & $0.94(\pm 0.38)$ & $0.79(\pm 0.12)$ & $0.98(\pm 0.15)$ & $3.44(\pm 0.67)$ \\
		
		\noalign{\smallskip}\hline
	\end{tabular}}
\end{table}

We also verify the effectiveness of temporal decomposition methods. In experiments 6-8, the decomposition operation is removed and just train 5 Transformers and 5 CNNs respectively. It turns out that the decomposition method boosts the forecasting accuracy significantly due to the design of decomposition kernels and Transformer and CNN networks. Putting this together justifies the choice on using $N=5$ kernels with temporal decomposition. It is also worth noting that the selection of $N$ is based on datasets while the proposed method easily supports different decomposition kernels, and rigorous experiments such as calculating similarity via \eqref{equ:similarity} are required to acquire the best settings.

%From all the ablation experiment groups, we empirically give our explanations for select $N=5$ decomposition kernels. Given the memory constraints, we only carry out experiments up to $N=5$ kernels. Naturally, the more kernels we have, the more temporal features can be gained and thus we can have better interpretability. Accuracy is also one metric for interpretable models, thus a trade-off is often considered. In our ablation study, $N=4$ decomposition kernels only achieved a slightly better accuracy compared with $N=5$. In order to obtain better temporal interpretability, we still choose $N=5$ kernels. Besides, we also derived the importance of our temporal decomposition method empirically.

 %However, it is also noticeable that our method cannot provide a uniform framework for most datasets, which requires more rigorous experiments to figure out the best settings. 

\begin{table}[tp]
	\centering
	\caption{Perturbation Experiment}
	\label{tab: 5}  
	\scalebox{0.78}{
	\begin{tabular}{lllll}
		\hline\noalign{\smallskip}	
		Model & MSE & MAE & RMSE &MAPE(\%)  \\
		\noalign{\smallskip}\hline\noalign{\smallskip}
	    Proposed   & $0.52(\pm 0.20)$ & $0.57(\pm 0.13)$ & $0.72(\pm 0.15)$ & $\textbf{2.16}(\pm 0.36)$  \\
		  Perturbation 1 (w/o $F_5$) & $0.72(\pm 0.24)$ & $0.69(\pm 0.15)$ & $0.85(\pm 0.16)$ & $2.72(\pm 0.44)$ \\
		Perturbation 2 (w/o $F_3$, $F_5$) & $0.80(\pm 0.29)$ & $0.70(\pm 0.15)$  & $0.95(\pm 0.25)$ & $3.14(\pm 0.43)$ \\
            Perturbation 3 (w/o $T_2$) & $0.57(\pm 0.18)$ & $0.59(\pm 0.12)$ & $0.75(\pm 0.16)$ & $2.35(\pm 0.62)$  \\
            Perturbation 4 (w/o $T_2$, $R_2$) & $0.64(\pm 0.26)$ & $0.60(\pm 0.16)$ & $0.80(\pm 0.19)$ & $2.56(\pm 0.37)$ \\
		
		\noalign{\smallskip}\hline
	\end{tabular} }
\end{table}

A perturbation study to verify the effectiveness of the interpretability method is also carried out, which is similar to approahces described in \cite{ribeiro2016should, fong2017interpretable, koh2017understanding}. Since the heatmaps illustrate the significance scores of the temporal and auxiliary features to the model outputs, the important features are eliminated during training and the impacts to accuracy will be examined. By observing how the model's prediction changes when that particular feature is absent or neutralized, the importance of that feature for the given prediction can be further analyzed.
In the first perturbation experiment, the calendar feature is eliminated, which has the highest significance score. In the second experiment, the two most significant features are removed: calendar and humidity. Without the important auxiliary features, the outputs of the model have experienced a drop in accuracy. Similarly observations hold for Perturbation experiment 3 and 4, where decomposition kernel of 12 hours ($T_2$), and both $T_2$ and residual kernel $R_2$ are removed respectively.

\subsection{Generalization Study}
To test the generalization capability of the proposed model, extra test experiments are also carried out. In the first test experiment, the Belgium load data from 1st January 2020 to 31st December 2021 ($\text{Test}_1$) is collected. The significant scores of the auxiliary and temporal features can also be viewed in the second subfigure of Fig. \ref{Heat2} and Fig. \ref{Heat3}. Compared with the first subgraph, the significance scores in this test experiment illustrate high consistency. The similarities for the temporal and auxiliary features scores of these two datasets are nearly about 99\%. Still, calendar and humidity are the two most significant features. For temporal features, trend and residual features of 6, 12, and 24 hours are critical to both outputs. As in the other test experiment where the dataset is sampled from New South Wales of Australia ($\text{Test}_2$). \textcolor{blue}{Both $\text{Test}_1$ and $\text{Test}_2$ are different from the training data and will be trained separately}. It is obvious that the temporal and auxiliary features have illustrated very different patterns with the datasets from Belgium. 

\section{Conclusion}
\label{conclusion}
In this paper, an interpretable STLF method via a multi-scale temporal decomposition model is proposed. The proposed model provides both temporal and auxiliary forecasting feature interpretability in a simple manner. Compared with the common baseline methods, the proposed model has achieved both better accuracy and interpretability. \textcolor{blue}{The MSE, MAE, RMSE, and MAPE values of the proposed method are 0.52, 0.57, 0.72, and 2.16\% respectively.} To illustrate the interpretability, we utilize heatmaps to denote the significance scores of different temporal auxiliary features in the model outputs. Besides, the proposed model can adapt to different datasets for its flexible settings. \textcolor{blue}{We note that the proposed method assumes that the forecasting target exhibits trend and seasonal patterns to decompose, and may not give good interpretations under scenarios with weaker patterns. In the  future work we are interested in automatic parameter tuning study that can figure out the best decomposition kernel size and the number of decomposition kernels, thus the manual finetuning process can be saved. }

% The 'triggered' command can be changed if desired:


%\bibliographystyle{IEEEtran}
\bibliographystyle{IEEEtran}
\bibliography{references}

\begin{thebibliography}{10}
\providecommand{\url}[1]{#1}
\csname url@samestyle\endcsname
\providecommand{\newblock}{\relax}
\providecommand{\bibinfo}[2]{#2}
\providecommand{\BIBentrySTDinterwordspacing}{\spaceskip=0pt\relax}
\providecommand{\BIBentryALTinterwordstretchfactor}{4}
\providecommand{\BIBentryALTinterwordspacing}{\spaceskip=\fontdimen2\font plus
\BIBentryALTinterwordstretchfactor\fontdimen3\font minus
  \fontdimen4\font\relax}
\providecommand{\BIBforeignlanguage}[2]{{%
\expandafter\ifx\csname l@#1\endcsname\relax
\typeout{** WARNING: IEEEtran.bst: No hyphenation pattern has been}%
\typeout{** loaded for the language `#1'. Using the pattern for}%
\typeout{** the default language instead.}%
\else
\language=\csname l@#1\endcsname
\fi
#2}}
\providecommand{\BIBdecl}{\relax}
\BIBdecl

\bibitem{gross1987short}
G.~Gross and F.~D. Galiana, ``Short-term load forecasting,'' \emph{Proceedings
  of the IEEE}, vol.~75, no.~12, pp. 1558--1573, 1987.

\bibitem{castelvecchi2016can}
D.~Castelvecchi, ``Can we open the black box of ai?'' \emph{Nature News}, vol.
  538, no. 7623, p.~20, 2016.

\bibitem{chen2018machine}
Y.~Chen, Y.~Tan, and D.~Deka, ``Is machine learning in power systems
  vulnerable?'' in \emph{2018 IEEE International Conference on Communications,
  Control, and Computing Technologies for Smart Grids (SmartGridComm)}.\hskip
  1em plus 0.5em minus 0.4em\relax IEEE, 2018, pp. 1--6.

\bibitem{chen2019exploiting}
Y.~Chen, Y.~Tan, and B.~Zhang, ``Exploiting vulnerabilities of load forecasting
  through adversarial attacks,'' in \emph{Proceedings of the tenth ACM
  international conference on future energy systems}, 2019, pp. 1--11.

\bibitem{zhou2022robust}
Y.~Zhou, Z.~Ding, Q.~Wen, and Y.~Wang, ``Robust load forecasting towards
  adversarial attacks via bayesian learning,'' \emph{IEEE Transactions on Power
  Systems}, vol.~38, no.~2, pp. 1445--1459, 2022.

\bibitem{liu2022learning}
R.~Liu and Y.~Chen, ``Learning task-aware energy disaggregation: a federated
  approach,'' in \emph{2022 IEEE 61st Conference on Decision and Control
  (CDC)}.\hskip 1em plus 0.5em minus 0.4em\relax IEEE, 2022, pp. 4412--4418.

\bibitem{almeshaiei2011methodology}
E.~Almeshaiei and H.~Soltan, ``A methodology for electric power load
  forecasting,'' \emph{Alexandria Engineering Journal}, vol.~50, no.~2, pp.
  137--144, 2011.

\bibitem{rojat2021explainable}
T.~Rojat, R.~Puget, D.~Filliat, J.~Del~Ser, R.~Gelin, and
  N.~D{\'\i}az-Rodr{\'\i}guez, ``Explainable artificial intelligence (xai) on
  timeseries data: A survey,'' \emph{arXiv preprint arXiv:2104.00950}, 2021.

\bibitem{jain2019attention}
S.~Jain and B.~C. Wallace, ``Attention is not explanation,'' \emph{arXiv
  preprint arXiv:1902.10186}, 2019.

\bibitem{gurses2022introducing}
G.~G{\"u}rses-Tran, T.~A. K{\"o}rner, and A.~Monti, ``Introducing
  explainability in sequence-to-sequence learning for short-term load
  forecasting,'' \emph{Electric Power Systems Research}, vol. 212, p. 108366,
  2022.

\bibitem{burkart2021survey}
N.~Burkart and M.~F. Huber, ``A survey on the explainability of supervised
  machine learning,'' \emph{Journal of Artificial Intelligence Research},
  vol.~70, pp. 245--317, 2021.

\bibitem{brockwell2002introduction}
P.~J. Brockwell and R.~A. Davis, \emph{Introduction to time series and
  forecasting}.\hskip 1em plus 0.5em minus 0.4em\relax Springer, 2002.

\bibitem{enders2008applied}
W.~Enders, \emph{Applied econometric time series}.\hskip 1em plus 0.5em minus
  0.4em\relax John Wiley \& Sons, 2008.

\bibitem{chang2021node}
C.-H. Chang, R.~Caruana, and A.~Goldenberg, ``Node-gam: Neural generalized
  additive model for interpretable deep learning,'' \emph{arXiv preprint
  arXiv:2106.01613}, 2021.

\bibitem{obst2021adaptive}
D.~Obst, J.~De~Vilmarest, and Y.~Goude, ``Adaptive methods for short-term
  electricity load forecasting during covid-19 lockdown in france,'' \emph{IEEE
  transactions on power systems}, vol.~36, no.~5, pp. 4754--4763, 2021.

\bibitem{agarwal2021neural}
R.~Agarwal, L.~Melnick, N.~Frosst, X.~Zhang, B.~Lengerich, R.~Caruana, and
  G.~E. Hinton, ``Neural additive models: Interpretable machine learning with
  neural nets,'' \emph{Advances in neural information processing systems},
  vol.~34, pp. 4699--4711, 2021.

\bibitem{zhang2023accurate}
M.~Zhang, Y.~Han, A.~S. Zalhaf, C.~Wang, P.~Yang, C.~Wang, S.~Zhou, and
  T.~Xiong, ``Accurate ultra-short-term load forecasting based on load
  characteristic decomposition and convolutional neural network with
  bidirectional long short-term memory model,'' \emph{Sustainable Energy, Grids
  and Networks}, vol.~35, p. 101129, 2023.

\bibitem{han2022efficient}
Y.~Han, Y.~Feng, P.~Yang, L.~Xu, and A.~S. Zalhaf, ``An efficient algorithm for
  atomic decomposition of power quality disturbance signals using convolutional
  neural network,'' \emph{Electric Power Systems Research}, vol. 206, p.
  107790, 2022.

\bibitem{vaswani2017attention}
A.~Vaswani, N.~Shazeer, N.~Parmar, J.~Uszkoreit, L.~Jones, A.~N. Gomez,
  {\L}.~Kaiser, and I.~Polosukhin, ``Attention is all you need,''
  \emph{Advances in neural information processing systems}, vol.~30, 2017.

\bibitem{li2020exploring}
C.~Li, Z.~Bao, L.~Li, and Z.~Zhao, ``Exploring temporal representations by
  leveraging attention-based bidirectional lstm-rnns for multi-modal emotion
  recognition,'' \emph{Information Processing \& Management}, vol.~57, no.~3,
  p. 102185, 2020.

\bibitem{hermann2020shapes}
K.~Hermann and A.~Lampinen, ``What shapes feature representations? exploring
  datasets, architectures, and training,'' \emph{Advances in Neural Information
  Processing Systems}, vol.~33, pp. 9995--10\,006, 2020.

\bibitem{olah2017feature}
C.~Olah, A.~Mordvintsev, and L.~Schubert, ``Feature visualization,''
  \emph{Distill}, vol.~2, no.~11, p.~e7, 2017.

\bibitem{zeng2023transformers}
A.~Zeng, M.~Chen, L.~Zhang, and Q.~Xu, ``Are transformers effective for time
  series forecasting?'' in \emph{Proceedings of the AAAI conference on
  artificial intelligence}, vol.~37, no.~9, 2023, pp. 11\,121--11\,128.

\bibitem{zhou2022fedformer}
T.~Zhou, Z.~Ma, Q.~Wen, X.~Wang, L.~Sun, and R.~Jin, ``Fedformer: Frequency
  enhanced decomposed transformer for long-term series forecasting,'' in
  \emph{International Conference on Machine Learning}.\hskip 1em plus 0.5em
  minus 0.4em\relax PMLR, 2022, pp. 27\,268--27\,286.

\bibitem{wu2021autoformer}
H.~Wu, J.~Xu, J.~Wang, and M.~Long, ``Autoformer: Decomposition transformers
  with auto-correlation for long-term series forecasting,'' \emph{Advances in
  Neural Information Processing Systems}, vol.~34, pp. 22\,419--22\,430, 2021.

\bibitem{daubechies1990wavelet}
I.~Daubechies, ``The wavelet transform, time-frequency localization and signal
  analysis,'' \emph{IEEE transactions on information theory}, vol.~36, no.~5,
  pp. 961--1005, 1990.

\bibitem{park2022vision}
N.~Park and S.~Kim, ``How do vision transformers work?'' \emph{arXiv preprint
  arXiv:2202.06709}, 2022.

\bibitem{sundararajan2020many}
M.~Sundararajan and A.~Najmi, ``The many shapley values for model
  explanation,'' in \emph{International conference on machine learning}.\hskip
  1em plus 0.5em minus 0.4em\relax PMLR, 2020, pp. 9269--9278.

\bibitem{garreau2020explaining}
D.~Garreau and U.~Luxburg, ``Explaining the explainer: A first theoretical
  analysis of lime,'' in \emph{International conference on artificial
  intelligence and statistics}.\hskip 1em plus 0.5em minus 0.4em\relax PMLR,
  2020, pp. 1287--1296.

\bibitem{karim2017lstm}
F.~Karim, S.~Majumdar, H.~Darabi, and S.~Chen, ``Lstm fully convolutional
  networks for time series classification,'' \emph{IEEE access}, vol.~6, pp.
  1662--1669, 2017.

\bibitem{ribeiro2016should}
M.~T. Ribeiro, S.~Singh, and C.~Guestrin, ``" why should i trust you?"
  explaining the predictions of any classifier,'' in \emph{Proceedings of the
  22nd ACM SIGKDD international conference on knowledge discovery and data
  mining}, 2016, pp. 1135--1144.

\bibitem{fong2017interpretable}
R.~C. Fong and A.~Vedaldi, ``Interpretable explanations of black boxes by
  meaningful perturbation,'' in \emph{Proceedings of the IEEE international
  conference on computer vision}, 2017, pp. 3429--3437.

\bibitem{koh2017understanding}
P.~W. Koh and P.~Liang, ``Understanding black-box predictions via influence
  functions,'' in \emph{International conference on machine learning}.\hskip
  1em plus 0.5em minus 0.4em\relax PMLR, 2017, pp. 1885--1894.

\end{thebibliography}

\end{document}